\documentclass[11pt,a4paper]{article}
\usepackage{acl2018}
\usepackage{times}
\usepackage{latexsym}
\usepackage{amsmath}
\usepackage{amsfonts,amssymb}
\usepackage{url}
\usepackage{graphicx}
\usepackage{subcaption}
\usepackage{enumitem}
\usepackage{tabularx}
\usepackage{multirow}

\aclfinalcopy 

\title{Large-Scale QA-SRL Parsing}

\author{
 Nicholas FitzGerald\thanks{~~Much of this work was done while these authors were at the Allen Institute for Artificial Intelligence.}
 \qquad Julian Michael$^*$
 \qquad Luheng He
 \qquad Luke Zettlemoyer$^*$\\
Paul G. Allen School of Computer Science and Engineering \\
University of Washington, Seattle, WA\\
\texttt{\{nfitz,julianjm,luheng,lsz\}@cs.washington.edu}
}

\date{}

\begin{document}
\maketitle
\begin{abstract}
    We present a new large-scale corpus of Question-Answer driven Semantic Role 
    Labeling (QA-SRL) annotations,
    and the first high-quality QA-SRL parser.
    Our corpus, QA-SRL Bank 2.0, consists of over 250,000 question-answer pairs for over 64,000 sentences across 3 domains and was gathered with a new crowd-sourcing scheme that we show has high precision and good recall at modest cost. 
    We also present neural models for two QA-SRL subtasks: detecting argument spans for a predicate and generating questions to label the semantic relationship.
    The best models achieve question accuracy of 82.6\% and span-level accuracy of 77.6\% (under human evaluation) on the full pipelined QA-SRL prediction task. They can also, as we show, be used to gather additional annotations at low cost. 
    
  
\end{abstract}

\section{Introduction}

\begin{figure}
    \centering
    \includegraphics[width=\columnwidth]{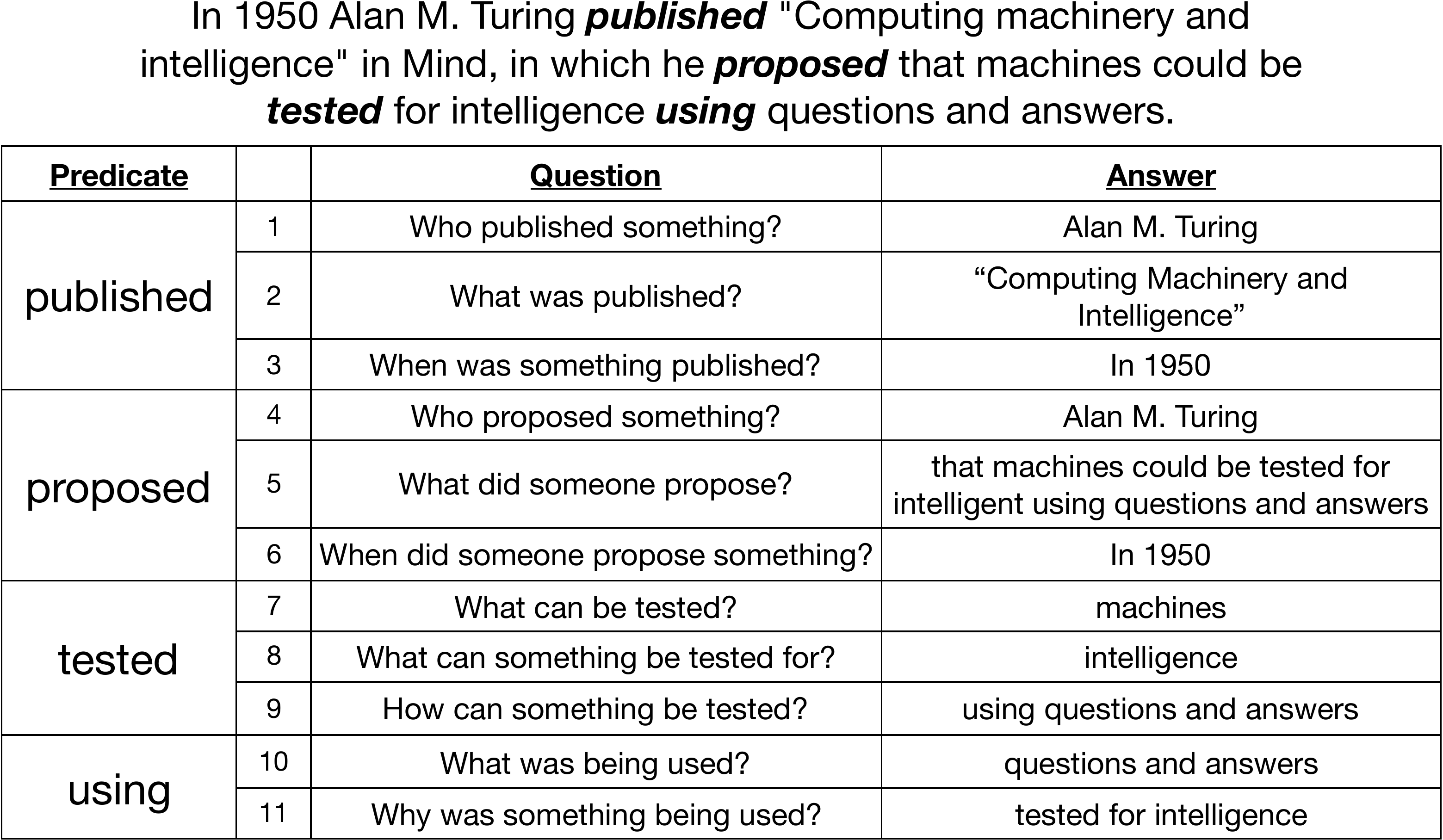}
    \vspace{-20pt}
    \setlength{\belowcaptionskip}{-15pt}
    \caption{An annotated sentence from our dataset. Question 6 was not produced by crowd workers in the initial collection, but was produced by our parser as part of Data Expansion (see~\autoref{sec:expansion}.)}
    \label{fig:data_example}
\end{figure}

Learning semantic parsers to predict the predicate-argument structures of a sentence is a long standing, open challenge~\citep{palmer2005proposition,baker1998berkeley}. Such systems are typically trained from datasets that are difficult to gather,\footnote{The PropBank~\citep{bonial2010propbank} and FrameNet~\citep{ruppenhofer2016framenet} annotation guides are 89 and 119 pages, respectively.} but recent research has explored training non-experts to provide this style of semantic supervision~\citep{abend2013universal,basile2012developing,reisinger2015semantic,he2015qasrl}. In this paper, we show for the first time that it is possible to go even further by crowdsourcing a large scale dataset that can be used to train high quality parsers at modest cost.   

We adopt the Question-Answer-driven Semantic Role Labeling  (QA-SRL)~\cite{he2015qasrl} annotation scheme. QA-SRL is appealing because it is intuitive to non-experts, has been shown to closely match the structure of traditional predicate-argument structure annotation schemes~\cite{he2015qasrl}, and has been used for end tasks such as Open IE~\cite{stanovsky2016emnlp}.
In QA-SRL, each predicate-argument relationship is labeled with a question-answer pair (see~\autoref{fig:data_example}).
\newcite{he2015qasrl} showed that high precision QA-SRL annotations can be gathered with limited training but that high recall is 
challenging to achieve; it is relatively easy to gather answerable questions, but difficult to ensure that every possible question is labeled for every verb. 
For this reason, they hired and trained hourly annotators and only labeled a relatively small dataset (3000 sentences). 

Our first contribution is a new, scalable approach for crowdsourcing QA-SRL. We introduce a streamlined web interface (including an auto-suggest mechanism and automatic quality control to boost recall) and use a validation stage to ensure high precision (i.e. all the questions must be answerable).
With this approach, we produce QA-SRL Bank 2.0, a dataset with 133,479 verbs from 64,018 sentences across 3 domains, totaling 265,140 question-answer pairs, in just 9 days. Our analysis shows that the data has high precision with good recall, although it does not cover every possible question. Figure~\ref{fig:data_example} shows example annotations.

Using this data, our second contribution is a comparison of several new models for learning a QA-SRL parser. We follow a pipeline approach where the parser does
(1) unlabeled {\em span detection} to determine the arguments of a given verb, and (2) {\em question generation} to label the relationship between the predicate
and each detected span.
Our best model uses a span-based representation similar to that introduced by~\newcite{lee2016learning} and a custom LSTM to decode questions from a learned span encoding. 
Our model does not require syntactic information and can be trained directly from the crowdsourced span labels.

Experiments demonstrate that the model does well on our new data, achieving up to 82.2\% span-detection F1 and 47.2\% exact-match question accuracy relative to the human annotations.
We also demonstrate the utility of learning to predict easily interpretable QA-SRL structures, using a simple data bootstrapping approach to expand our dataset further.
By tuning our model to favor recall, we over-generate questions which can be validated using our annotation pipeline, 
allowing for greater recall without requiring costly redundant annotations in the question writing step.
Performing this procedure on the training and development sets grows them by 20\% and leads to improvements when retraining our models. 
Our final parser is highly accurate, achieving 82.6\% question accuracy and 77.6\% span-level precision in an human evaluation.
Our data, code, and trained models will be made publicly available.\footnote{http://qasrl.org}
\section{Data Annotation}
\label{sec:data}

\begin{table*}
\small
\centering
\begin{tabular}{lllllll}
    \textbf{Wh} & \textbf{Aux} & \textbf{Subj} & \textbf{Verb} & \textbf{Obj} & \textbf{Prep} & \textbf{Misc} \\
    Who & & & blamed & someone & & \\
    What & did & someone & blame & something & on & \\
    Who & & & refused & & to & do something \\
    When & did & someone & refuse & & to & do something \\
    Who & might & & put & something & & somewhere \\
    Where & might & someone & put & something & &
\end{tabular}
\caption{Example QA-SRL questions, decomposed into their slot-based representation. See~\newcite{he2015qasrl} for the full details. All slots draw from a small, deterministic set of options, including verb tense ($present$, $pastparticiple$, etc.) Here we have replaced the verb-tense slot with its conjugated form.}
\label{tab:qasrl-format}
\end{table*}

A QA-SRL annotation consists of a set of question-answer pairs for each verbal predicate in a sentence, where each answer is a set of contiguous spans from the sentence.
QA-SRL questions are defined by a 7-slot template shown in~\autoref{tab:qasrl-format}.
We introduce a crowdsourcing pipeline to collect annotations rapidly, cheaply, and at large scale.

\paragraph{Pipeline}
Our crowdsourcing pipeline consists of a \textit{generation} and \textit{validation} step. In the generation step, a sentence with one of its verbs marked is shown
to a single worker, who must write QA-SRL questions for the verb and highlight their answers in the sentence. The questions are passed to the validation step,
where \( n \) workers answer each question or mark it as \textit{invalid}. In each step, no two answers to distinct questions may overlap with each other, to prevent redundancy.

\paragraph{Instructions}
Workers are instructed that a \textit{valid} question-answer pair must satisfy three criteria: 1) the question is grammatical, 2) the question-answer pair is asking about the time, place, participants, etc., of the target verb, and 3) all correct answers to each question are given.

\begin{figure}
    \centering
    \includegraphics[width=\columnwidth]{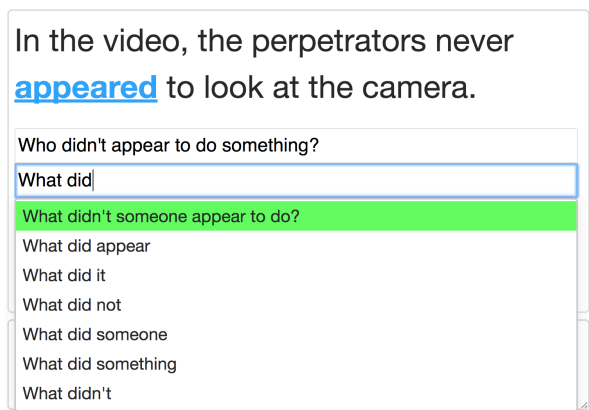}
    \vspace{-20pt}
    \setlength{\belowcaptionskip}{-10pt}
    \caption{Interface for the generation step. Autocomplete shows completions of the current QA-SRL slot, and auto-suggest shows fully-formed questions (highlighted green) based on the previous questions. }
    \label{fig:qasrl-interface}
\end{figure}

\paragraph{Autocomplete}
We provide an autocomplete drop-down to streamline question writing.
Autocomplete is implemented as a Non-deterministic Finite Automaton (NFA) whose states correspond to the 7 QA-SRL slots paired with a partial representation of the question's syntax.
We use the NFA to make the menu more compact by disallowing obviously ungrammatical combinations (e.g., \textit{What did been appeared?}), and the syntactic representation to 
auto-suggest complete questions about arguments that have not yet been covered (see \autoref{fig:qasrl-interface}).
The auto-suggest feature significantly reduces the number of keystrokes required to enter new questions after the first one, speeding up the annotation process and making it easier for annotators to provide higher recall.

\paragraph{Payment and quality control}
Generation pays 5c for the first QA pair (required), plus 5c, 6c, etc. for each successive QA pair (optional), to boost recall. The validation step pays 8c per verb, plus a 2c bonus per question beyond four. Generation workers must write at least 2 questions per verb and have 85\% of their questions counted valid, and validators must maintain 85\% answer span agreement with others, or they are disqualified from further work.
A validator's answer is considered to agree with others if their answer span overlaps with answer spans provided by a majority of workers.

\paragraph{Preprocessing}
We use the Stanford CoreNLP tools \cite{manning2014stanford} for sentence segmentation, tokenizing, and POS-tagging.
We identify verbs by POS tag, with heuristics to filter out auxiliary verbs while retaining non-auxiliary uses of ``have'' and ``do.''
We identify conjugated forms of each verb for the QA-SRL templates by finding them in Wiktionary.\footnote{www.wiktionary.org}

\paragraph{Dataset}

\begin{table}
\centering
\small
\begin{tabular}{r|r|r|r}
                       & \textbf{Wikipedia}    & \textbf{Wikinews} & \textbf{Science}  \\ \hline
   \textbf{Sentences}  & 15,000                & 14,682            & 46,715            \\
   \textbf{Verbs}      & 32,758                & 34,026            & 66,653            \\
   \textbf{Questions}  & 75,867                & 80,081            & 143,388           \\
   \textbf{Valid Qs}   & 67,146                & 70,555            & 127,455
\end{tabular}
\caption{Statistics for the dataset with questions written by workers across three domains.}
\label{tab:dataset-stats}
\end{table}

We gathered annotations for 133,479 verb mentions in 64,018 sentences (1.27M tokens) across 3 domains:
Wikipedia,
Wikinews,
and science textbook text from the Textbook Question Answering (TQA) dataset~\cite{kembhavi2017tqa}. 
We partitioned the source documents into train, dev, and test, sampled paragraph-wise from each document with an 80/10/10 split by sentence.

Annotation in our pipeline with \(n = 2\) validators took 9 days on Amazon Mechanical Turk.\footnote{www.mturk.com}
1,165 unique workers participated, annotating a total of 299,308 questions.
Of these, 265,140 (or 89\%) were considered valid by both validators,
for an average of 1.99 valid questions per verb and 4.14 valid questions per sentence.
See \autoref{tab:dataset-stats} for a breakdown of dataset statistics by domain.
The total cost was \$43,647.33, for an average of 32.7c per verb mention, 14.6c per question, or 16.5c per valid question.
For comparison, \newcite{he2015qasrl} interviewed and hired contractors to annotate data at much smaller scale for a cost of about 50c per verb. Our annotation scheme is cheaper, far more scalable, and provides more (though noisier) supervision for answer spans.

To allow for more careful evaluation, we  validated 5,205 sentences at a higher density (up to 1,000 for each domain in dev and test),
re-running the generated questions through validation with \(n = 3\) for a total of 6 answer annotations for each question.

\paragraph{Quality}
Judgments of question validity had moderate agreement. About 89.5\% of validator judgments rated a question as valid, and the agreement rate between judgments of the same question on whether the question is invalid is 90.9\%. This gives a Fleiss's Kappa of 0.51.
In the higher-density re-run, validators were primed to be more critical: 76.5\% of judgments considered a question valid, and agreement was at 83.7\%, giving a Fleiss's Kappa of 0.55.

Despite being more critical in the denser annotation round, questions marked valid in the original dataset were marked valid by the new annotators in 86\% of cases, showing our data's relatively high precision. The high precision of our annotation pipeline is also backed up by our small-scale manual evaluation (see Coverage below).

Answer spans for each question also exhibit good agreement. On the original dataset, each answer span has a 74.8\% chance to exactly match one provided by another annotator (up to two), and on the densely annotated subset, each answer span has an 83.1\% chance to exactly match one provided by another annotator (up to five). 

\paragraph{Coverage}
Accurately measuring recall for QA-SRL annotations is an open challenge.
For example, question 6 in \autoref{fig:data_example} reveals an inferred temporal relation that would not be annotated as part of traditional SRL.
Exhaustively enumerating the full set of such questions is difficult, even for experts.

However, we can compare to the original QA-SRL dataset~\cite{he2015qasrl}, where Wikipedia sentences were annotated with 2.43 questions per verb.
Our data has lower---but loosely comparable---recall, with 2.05 questions per verb in Wikipedia.

In order to further analyze the quality of our annotations relative to~\cite{he2015qasrl},
we reannotate a 100-verb subset of their data both manually (aiming for exhaustivity) and with our crowdsourcing pipeline.
We merge the three sets of annotations, manually remove bad questions (and their answers), and calculate the precision and recall of the crowdsourced annotations and those of \newcite{he2015qasrl} against this pooled, filtered dataset (using the span detection metrics described in \autoref{sec:initial-results}).
Results, shown in \autoref{tab:data_quality}, show that our pipeline produces comparable precision with only a modest decrease in recall.
Interestingly, re-adding the questions rejected in the validation step
greatly increases recall with only a small decrease in precision, showing that validators sometimes rejected questions considered valid by the authors.
However, we use the filtered dataset for our experiments, and in \autoref{sec:expansion}, we show how another crowdsourcing step can further improve recall.

\begin{table}
    \centering
    \begin{tabular}{|l|c|c|c|}
        \hline
         & P & R & F \\
         \hline
         \newcite{he2015qasrl} & 97.5 & 86.6 & 91.7 \\
         This work & 95.7 & 72.4 & 82.4 \\
         This work (unfiltered) & 94.9 & 85.4 & 89.9 \\
         \hline
    \end{tabular}
    \caption{Precision and recall of our annotation pipeline on a merged and validated subset of 100 verbs.
    The unfiltered number represents relaxing the restriction that none of 2 validators
    marked the question as invalid.}
    \label{tab:data_quality}
\end{table}
\section{Models}

Given a sentence $\boldsymbol{X} = {x_0, \dots, x_n}$, the goal of a QA-SRL parser is to produce a
set of tuples $(v_i, \boldsymbol{Q_i}, \mathcal{S}_i)$, where $v \in \{0, \dots, n\}$ is the index of 
a verbal predicate, $\boldsymbol{Q}_i$ is a question, and $\mathcal{S}_i \in \{(i, j) \mid i, j \in [0, n], j \geq i\}$
is a set of spans which are valid answers.
Our proposed parsers construct these tuples in a three-step pipeline:
\begin{enumerate}[noitemsep,nolistsep]
    \item{\em Verbal predicates} are identified using the same POS-tags and heuristics as in data collection (see Section~\ref{sec:data}).
    \item {\em Unlabeled span detection} selects a set $\mathcal{S}_v$ of spans as arguments for a given verb $v$.
    \item {\em Question generation} predicts a question for each span in $\mathcal{S}_v$. Spans are then grouped by question, giving each question a set of answers.
\end{enumerate}

We describe two models for unlabeled span detection in section~\ref{sec:span-detection},
followed by question generation in section~\ref{sec:question-model}.
All models are built on an LSTM encoding of the sentence.
Like~\newcite{he2017deep}, we start with an input 
$\boldsymbol{X}_v = \{\boldsymbol{x}_0 \dots \boldsymbol{x}_n\}$, where the representation
$\boldsymbol{x}_i$ at each time step 
is a concatenation of the token $w_i$'s embedding and an embedded binary feature $(i=v)$ which indicates
whether $w_i$ is the predicate under consideration.
We then compute the output representation
\( \boldsymbol{H}_{v} = \textsc{BiLSTM}(\boldsymbol{X}_v) \)
using a stacked alternating LSTM~\cite{zhou2015end} with highway connections~\citep{srivastava2015training} and 
recurrent dropout~\citep{gal2016theoretically}.
Since the span detection and question generation models both use an LSTM encoding, this component could in principle
be shared between them. However, in preliminary experiments we found that sharing hurt performance, so for
the remainder of this work each model is trained independently.

\subsection{Span Detection}
\label{sec:span-detection}

Given an encoded sentence $\boldsymbol{H}_v$, the goal of 
span detection is to select the spans $\mathcal{S}_v$ 
that correspond to arguments of the given predicate. We explore two models: a sequence-tagging  model with BIO encoding, and a span-based model which assigns a probability to every possible span.

\subsubsection{BIO Sequence Model}

Our BIO model predicts a set of spans via a sequence $\boldsymbol{y}$ where each $y_i \in \{\boldsymbol{B}, \boldsymbol{I}, \boldsymbol{O}\}$, representing a token at the beginning, interior, or outside of any span, respectively. 
Similar to \newcite{he2017deep}, we make independent predictions for each token at training time, and use Viterbi decoding to enforce hard BIO-constraints\footnote{E.g., an $I$-tag should only follow a $B$-tag.} at test time. 
The resulting sequences are in one-to-one correspondence with sets $\mathcal{S}_v$ of spans which are pairwise non-overlapping.
The locally-normalized BIO-tag distributions are computed from the BiLSTM outputs~$\boldsymbol{H}_v = \{\boldsymbol{h}_{v0}, \ldots, \boldsymbol{h}_{vn}\}$:
\begin{equation}
    p(y_t \mid \boldsymbol{x}) \propto exp(\boldsymbol{w}_{\text{tag}}^{\intercal}\textsc{MLP}(\boldsymbol{h}_{vt}) + \boldsymbol{b}_{\text{tag}})
\end{equation}

\subsubsection{Span-based Model}
\label{sec:span-model}

Our span-based model makes independent binary decisions for all $O(n^2)$ spans in the sentence.
Following~\newcite{lee2016learning}, the representation of a span $(i,j)$ is the concatenation of the BiLSTM output at each endpoint:
\begin{equation}
   \boldsymbol{s}_{vij} = [\boldsymbol{h}_{vi}, \boldsymbol{h}_{vj}].
\end{equation}
The probability that the span is an argument of predicate $v$ is computed by the sigmoid function:
\begin{equation}
    p(y_{ij} \mid \boldsymbol{X}_v) = \sigma(\boldsymbol{w}_{\text{span}}^{\intercal}  \textsc{MLP}(\boldsymbol{s}_{vij}) + \boldsymbol{b}_{\text{span}})
\end{equation}
At training time, we minimize the binary cross entropy summed over all $n^2$ possible spans, counting a span as a positive example if it appears as an answer to any question.

At test time, we choose a threshold $\tau$ and select every span that the model assigns probability greater than $\tau$, allowing us to trade off precision and recall.

\subsection{Question Generation}
\label{sec:question-model}
We introduce two question generation models. 
Given a span representation $\boldsymbol{s}_{vij}$ defined in~\autoref{sec:span-model}, our models generate questions by picking a word for each question slot (see~\autoref{sec:data}).
Each model calculates a joint distribution $p(\boldsymbol{y} \mid \boldsymbol{X}_v, \boldsymbol{s}_{vij})$ over values \(\boldsymbol{y} = (y_1,\dots,y_7)\) for the question slots given a span $\boldsymbol{s}_{vij}$,
and is trained to minimize the negative log-likelihood of gold slot values.

\subsubsection{Local Model}

The local model predicts the words for each slot independently:
\begin{equation}
    p(y_k \mid \boldsymbol{X}_v, \boldsymbol{s}_{vij} ) \propto \exp(\boldsymbol{w}^{\intercal}_{k} \textsc{MLP}(\boldsymbol{s}_{vij}) + \boldsymbol{b}_{k}).
\end{equation}

\subsubsection{Sequence Model}

The sequence model uses the machinery of an RNN to share information between slots.
At each slot $k$, we apply a multiple layers of LSTM cells:
\begin{equation}
    \boldsymbol{h}_{l,k}, \boldsymbol{c}_{l, k} = \textsc{LSTMCell}_{l,k}(\boldsymbol{h}_{l-1, k}, \boldsymbol{h}_{l, k-1}, \boldsymbol{c}_{l, k-1})
\end{equation}
where the initial input at each slot is a concatenation of the span representation and the embedding of the previous word of the question:
$\boldsymbol{h}_{0, k} = [\boldsymbol{s}_{vij}; \boldsymbol{y}_{k-1}]$. 
Since each question slot predicts from a different set of words, we found it beneficial to use separate weights for the LSTM cells at each slot $k$.
During training, we feed in the gold token at the previous slot, while at test time, we use the predicted token.
The output distribution at slot $k$ is computed via the final layers' output vector $\boldsymbol{h}_{Lk}$:
\begin{equation}
    p(y_k \mid \boldsymbol{X}_v, \boldsymbol{s}_{vij}) \propto \exp(\boldsymbol{w}^{\intercal}_{k}  \textsc{MLP}(\boldsymbol{h}_{Lk}) + \boldsymbol{b}_{k})
\end{equation}

\section{Initial Results}
\label{sec:initial-results}
Automatic evaluation for QA-SRL parsing presents multiple challenges.
In this section, we introduce automatic metrics that can help us compare models.
In~\autoref{sec:final-evaluation}, we will report human evaluation results for our final system.

\subsection{Span Detection}
\paragraph{Metrics}
We evaluate span detection using a modified notion of precision and recall.
We count predicted spans as correct if they match any of the labeled spans in the dataset.
Since each predicted span could potentially be a match to multiple questions
(due to overlapping annotations) we map each predicted span to one matching question in the 
way that maximizes measured recall using maximum bipartite matching.
We use both exact match and intersection-over-union (IOU) greater than 0.5 as matching criteria.

\paragraph{Results}
\autoref{tab:initial-span-results} shows span detection results on the development set.
We report results for the span-based models at two threshold values \(\tau\):~\(\tau = 0.5\), and \(\tau = \tau^*\) maximizing F1.
The span-based model significantly improves over the BIO model in both precision and recall, 
although the difference is less pronounced under IOU matching.

\begin{table}
    \centering
    \begin{tabular}{|c|c|c|c|}
        \hline
        \multicolumn{4}{|c|}{Exact Match} \\
        \hline
        & P & R & F \\
        \hline
         BIO &  69.0 & 75.9 & 72.2 \\
         Span ($\tau = 0.5$) & 81.7 & 80.9 & 81.3 \\
         Span ($\tau = \tau*$) & 80.0 & 84.7 & 82.2 \\
        \hline
        \multicolumn{4}{|c|}{IOU $\geq$ 0.5} \\
        \hline
        & P & R & F \\
        \hline
         BIO &  80.4 & 86.0 & 83.1 \\
         Span ($\tau = 0.5$) & 87.5 & 84.2 & 85.8 \\
         Span ($\tau = \tau*$) & 83.8 & 93.0 & 88.1 \\
         \hline
    \end{tabular}
    \caption{Results for Span Detection on the dense development dataset. Span detection results are given with the
    cutoff threshold $\tau$ at 0.5, and at the value which maximizes F-score. The top chart lists precision, recall
    and F-score with exact span match, while the bottom reports matches where the intersection over union (IOU) is $\geq 0.5$.}
    \label{tab:initial-span-results}
\end{table}

\subsection{Question Generation}

\paragraph{Metrics}
Like all generation tasks, evaluation metrics for question generation must contend with the fact that there are in general 
multiple possible valid questions for a given predicate-argument pair.
For instance, the question ``Who did someone blame something on?'' may be rephrased as
``Who was blamed for something?''
However, due to the constrained space of possible questions defined by QA-SRL's slot format, accuracy-based metrics
can still be informative.
In particular, we report the rate at which the predicted question exactly matches the gold question, as well as
a relaxed match where we only count the question word (WH), subject (SBJ), object (OBJ) and Miscellaneous (Misc)
slots (see~\autoref{tab:qasrl-format}).
Finally, we report average slot-level accuracy.

\paragraph{Results}
~\autoref{tab:initial-quesgen-results} shows the results for question generation on the development set.
The sequential model's exact match accuracy is significantly higher, while 
word-level accuracy
is roughly comparable, reflecting the fact that the local model learns the slot-level posteriors.

\begin{table}
    \centering
    \begin{tabular}{|c|c|c|c|}
        \hline
         & EM & PM & SA \\
         \hline
         Local & 44.2 & 62.0 & 83.2 \\
         Seq. & 47.2 & 62.3 & 82.9 \\
         \hline
    \end{tabular}
    \caption{Question Generation results on the dense development set. {\bf EM} - Exact Match accuracy, {\bf PM} - Partial Match Accuracy, {\bf SA} - Slot-level accuracy}
    \label{tab:initial-quesgen-results}
\end{table}

\subsection{Joint results}

\autoref{tab:initial-joint-results} shows precision and recall for joint span detection and question generation, using exact match for both.
This metric is exceedingly hard, but it shows that almost 40\% of predictions are exactly correct in both span and question.
In~\autoref{sec:final-evaluation}, we use human evaluation to get a more accurate assessment of our model's accuracy.

\begin{table}
    \begin{tabular}{|c|c|c|c|}
        \hline
         & P & R & F \\
         \hline
         Span + Local & 37.8 & 43.7 & 40.6 \\
         Span + Seq. ($\tau=0.5$) & 39.6 & 45.8 & 42.4 \\
         \hline
    \end{tabular}
    \caption{Joint span detection and question generation results on the dense development set, using exact-match for both spans and questions.}
    \label{tab:initial-joint-results}
\end{table}

\section{Data Expansion}
\label{sec:expansion}

Since our trained parser can produce full QA-SRL annotations, its predictions can be validated by the same process as in our original
annotation pipeline, allowing us to focus annotation efforts towards filling potential data gaps.

By detecting spans at a low probability cutoff, we over-generate QA pairs for already-annotated sentences.
Then, we filter out QA pairs whose answers overlap with answer spans in the existing annotations, or whose questions match existing questions.
What remains are candidate QA pairs which fill gaps in the original annotation.
We pass these questions to the validation step of our crowdsourcing pipeline with \(n = 3\) validators, resulting in new labels.

We run this process on the training and development partitions of our dataset.
For the development set, we use the trained model described in the previous section.
For the training set, we use a relaxed version of jackknifing, training 5 models over 5 different folds.
We generate 92,080 questions at a threshold of $\tau=0.2$. Since in this case many sentences have only one question, we restructure the pay to a 2c base rate with a 2c bonus per question after the first (still paying no less than 2c per question).

\paragraph{Data statistics}
46,017 (50\%) of questions run through the expansion step were considered valid by all three annotators.
In total, after filtering, the expansion step increased the number of valid questions in the train and dev partitions by 20\%. However, for evaluation, since our recall metric identifies a single question for each answer span (via bipartite matching), we filter out likely question paraphrases by removing questions in the expanded development set whose answer spans have two overlaps with the answer spans of one question in the original annotations.
After this filtering, the expanded development set we use for evaluation has 11.5\% more questions than the original development set.

The total cost including MTurk fees was \$8,210.66, for a cost of 8.9c per question, or 17.8c per valid question.
While the cost per valid question was comparable to the initial annotation, we gathered many more negative examples (which may serve useful in future work), and this method allowed us to focus on questions that were missed in the first round and improve the exhaustiveness of the annotation (whereas it is not obvious how to make fully crowdsourced annotation more exhaustive at a comparable cost per question).

\paragraph{Retrained model}

We retrained our final model on the training set extended with the new valid questions,
yielding modest improvements on both span detection and question generation in the development set (see \autoref{tab:expansion-result}).
The span detection numbers are higher than on the original dataset, because the expanded development data captures true positives produced by the original model (and the resulting increase in precision can be traded off for recall as well).

\begin{table}
    \centering
    \begin{subtable}[t]{\columnwidth}
    \centering
    \begin{tabular}{|c|c|c|c|c|}
        \hline
        \multicolumn{5}{|c|}{Exact Match} \\
        \hline
        & P & R & F & AUC \\
        \hline
         Original & 80.8 & 86.8 & 83.7 & .906 \\
         Expanded & 82.9 & 86.4 & 84.6 & .910 \\
        \hline
        \multicolumn{5}{|c|}{IOU $\geq$ 0.5} \\
        \hline
        & P & R & F & AUC \\
        \hline
         Original & 87.1 & 93.2 & 90.1 & .946 \\
         Expanded & 87.9 & 93.1 & 90.5 & .949 \\
         \hline
    \end{tabular}
    \caption{Span Detection results with $\tau*$.}
    \end{subtable}
    \begin{subtable}[b]{\columnwidth}
    \centering
        \begin{tabular}{|c|c|c|c|}
        \hline
         & EM & PM & WA \\
         \hline
         Original & 50.5 & 64.4 & 84.1 \\
         Expanded & 50.8 & 64.9 & 84.1 \\
         \hline
    \end{tabular}
    \caption{Question Generation results}
    \end{subtable}
    \begin{subtable}[b]{\columnwidth}
    \centering
    \begin{tabular}{|c|c|c|c|}
        \hline
         & P & R & F \\
         \hline
         Original & 47.5 & 46.9 & 47.2 \\
         Expanded & 44.3 & 55.0 & 49.1 \\
         \hline
    \end{tabular}
    \caption{Joint span detection and question generation results with $\tau=0.5$}
    \end{subtable}
    \caption{Results on the expanded development set comparing the full model trained on the original data, and with the expanded data.}
    \label{tab:expansion-result}
\end{table}

\begin{figure*}[ht]
    \centering
    \begin{subfigure}[t]{\columnwidth}
        \centering
        \includegraphics[width=\columnwidth]{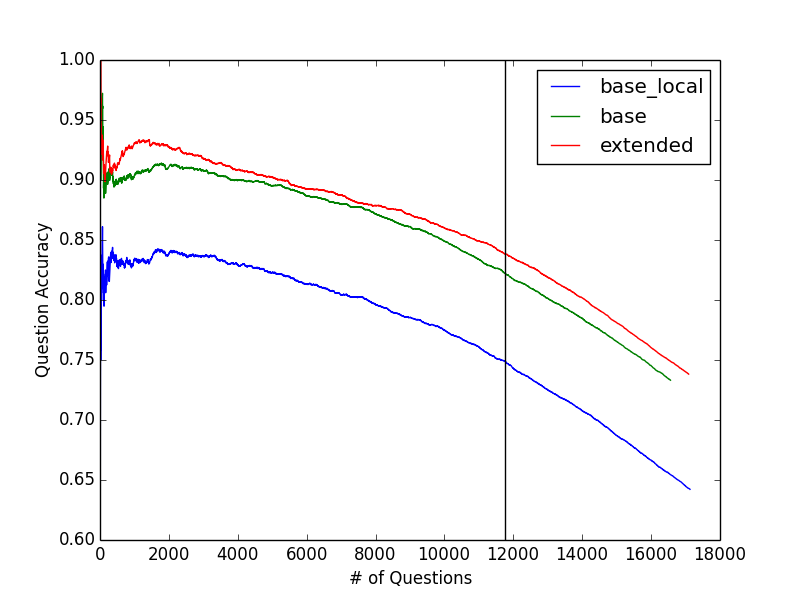}
        \caption{Question accuracy on Dev}
    \end{subfigure}
    \begin{subfigure}[t]{\columnwidth}
        \centering
        \includegraphics[width=\columnwidth]{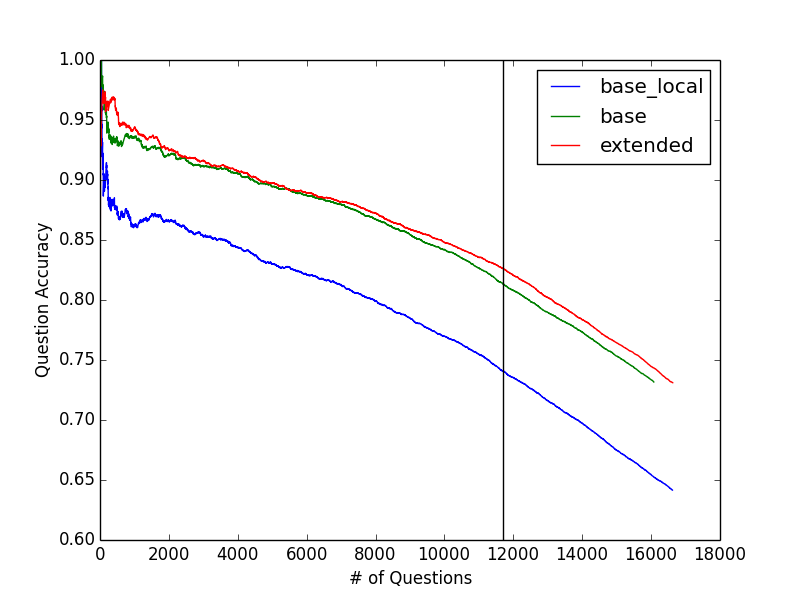}
        \caption{Question accuracy on Test}
    \end{subfigure}
    
    \begin{subfigure}[b]{\columnwidth}
        \centering
        \includegraphics[width=\columnwidth]{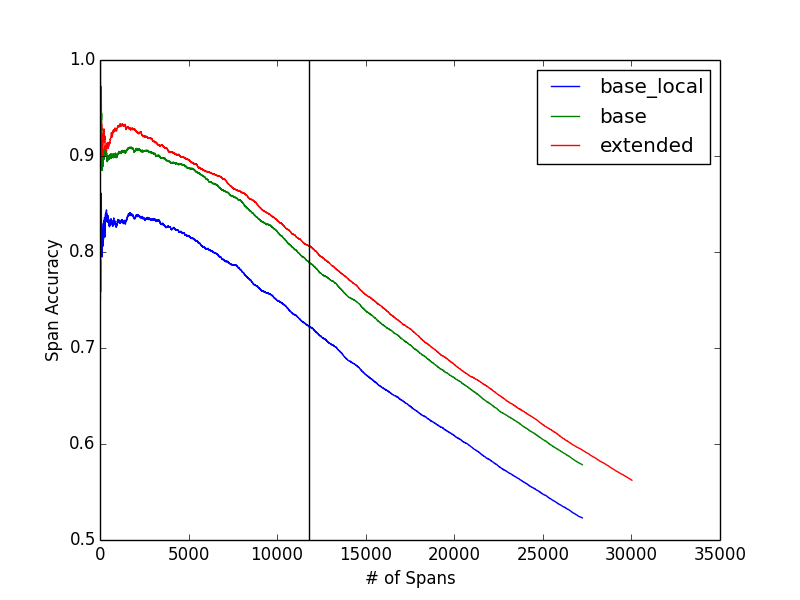}
        \caption{Span accuracy on Dev}
    \end{subfigure}
    \begin{subfigure}[b]{\columnwidth}
        \centering
        \includegraphics[width=\columnwidth]{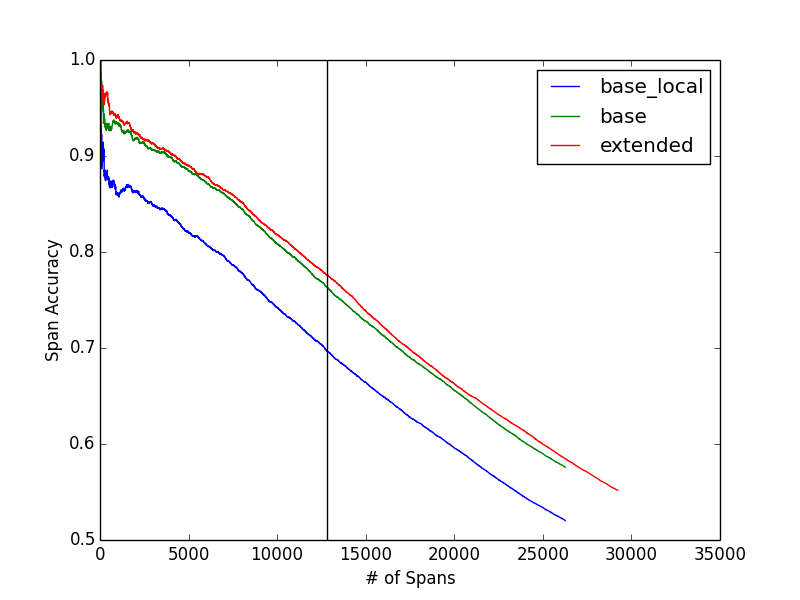}
        \caption{Span accuracy on Test}
    \end{subfigure}
    \caption{Human evaluation accuracy for questions and spans, as each model's span detection threshold is varied. Questions 
    are considered correct if 5 out of 6 annotators consider it valid. Spans are considered correct if their question
    was valid, and the span was among those labeled by human annotators for that question. The vertical line indicates a threshold
    value where the number of questions per sentence matches that of the original labeled data (2 questions / verb).}
    \label{fig:human-eval}
\end{figure*}

\section{Final Evaluation}
\label{sec:final-evaluation}

\begin{table*}
    \centering
    \includegraphics[width=0.8\textwidth]{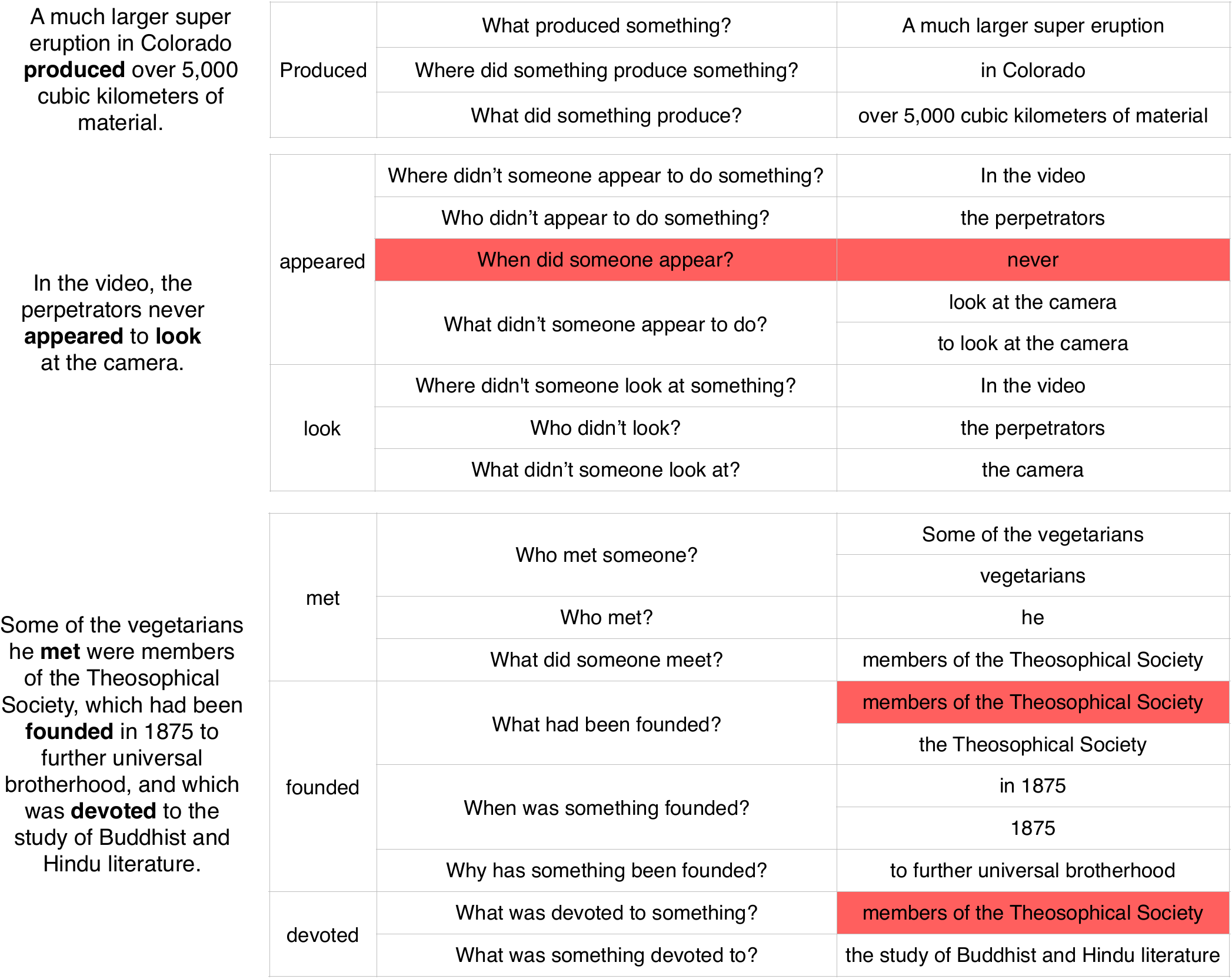}
    \caption{System output on 3 randomly sampled sentences from the development set (1 from each of the 3 domains). Spans were selected with $\tau = 0.5$. Questions
    and spans with a red background were marked incorrect during human evaluation.}
    \label{fig:output_example}
\end{table*}

We use the crowdsourced validation step to do a final human evaluation of our models.
We test 3 parsers: the span-based span detection model paired with each of the local and sequential question generation models trained on the
initial dataset, and our final model (span-based span detection and sequential question generation) trained with the expanded data.

\paragraph{Methodology}
On the 5,205 sentence densely annotated subset of dev and test, we generate QA-SRL labels with all of the models using a span detection threshold of $\tau=0.2$ and combine the questions with the existing data. We filter out questions that fail the autocomplete grammaticality check (counting them invalid) and pass the data into the validation step, annotating each question to a total of 6 validator judgments.
We then compute question and span accuracy as follows:
A question is considered correct if 5 out of 6 annotators consider it valid, and
a span is considered correct if its generated question is correct and the span is among those selected for the question by validators.
We rank all questions and spans by the threshold at which they are generated, which allows us to
compute accuracy at different levels of recall.

\paragraph{Results}
\autoref{fig:human-eval} shows the results.
As expected, the sequence-based question generation models are much more accurate than the local model; this is largely because the local model generated many questions that failed the grammaticality check.
Furthermore, training with our expanded data results in more questions and spans generated at the same threshold.
If we choose a threshold value which gives a similar number of questions per sentence as were labeled in the original data annotation (2 questions / verb), question and span accuracy are 82.64\% and 77.61\%, respectively.

~\autoref{fig:output_example} shows the output of our best system on 3 randomly selected sentences from our development set (one from each domain).
The model was overall highly accurate---only one question and 3 spans are considered incorrect, and each mistake is nearly correct,\footnote{The incorrect
question ``When did someone appear?'' would be correct if the Prep and Misc slots were corrected to read ``When did someone appear to do something?''}
even when the sentence contains a negation.

\section{Related Work}

Resources and formalisms for semantics often require expert annotation and underlying syntax \cite{palmer2005proposition,baker1998berkeley, banarescu2013abstract}. 
Some more recent semantic resources require less annotator training, or can be  crowdsourced \cite{abend2013universal,reisinger2015semantic,basile2012developing, michael2017qamr}. In particular, the original QA-SRL \cite{he2015qasrl} dataset is annotated by freelancers, while we developed streamlined crowdsourcing approaches for more scalable annotation.

Crowdsourcing has also been used for indirectly annotating syntax \cite{he2016human,duan2016generating}, and to complement expert annotation of SRL~\cite{wang2017crowd}. 
Our crowdsourcing approach draws heavily on that of~\newcite{michael2017qamr},
with automatic two-stage validation for the collected question-answer pairs.

More recently, models have been developed for these newer semantic resources, such as UCCA \cite{teichert2017semantic} and Semantic Proto-Roles \cite{white2017semantic}. Our work is the first high-quality parser for QA-SRL, which has several unique modeling challenges, such as its highly structured nature and the noise in crowdsourcing. 

Several recent works have explored neural models
for SRL tasks~\cite{collobert2007fast,fitzgerald2015semantic,swayamdipta2017frame,yang2017joint},
many of which employ a BIO encoding~\cite{zhou2015end,he2017deep}.
Recently, span-based models have proven to be useful for question answering~\cite{lee2016learning} and coreference resolution~\cite{lee2017end}, and PropBank SRL~\cite{he2018jointly}.
\section{Conclusion}

In this paper, we demonstrated that QA-SRL can be scaled to large datasets, enabling a new methodology for labeling
and producing predicate-argument structures at a large scale.
We presented a new, scalable approach for crowdsourcing QA-SRL, which allowed us to collect QA-SRL Bank 2.0, a new dataset covering
over 250,000 question-answer pairs from over 64,000 sentences, in just 9 days.
We demonstrated the utility of this data by training the first parser which is able to produce high-quality 
QA-SRL structures.
Finally, we demonstrated that the validation stage of our crowdsourcing pipeline, in combination with our parser
tuned for recall, can be used to add new annotations to the dataset, increasing recall.
\section*{Acknowledgements}

The crowdsourcing funds for QA-SRL Bank 2.0 was provided by the Allen Institute for Artificial Intelligence.
This research was supported in part by the ARO (W911NF-16-1-0121) the NSF (IIS-1252835, IIS-1562364), a gift from Amazon, and an Allen Distinguished Investigator Award.
We would like to thank Gabriel Stanovsky and Mark Yatskar for their helpful feedback.

\bibliographystyle{acl_natbib}
\bibliography{acl2018}

\end{document}